\documentclass{article}

\usepackage{PRIMEarxiv}

\usepackage[utf8]{inputenc} % allow utf-8 input
\usepackage[T1]{fontenc}    % use 8-bit T1 fonts
\usepackage{hyperref}       % hyperlinks
\usepackage{url}            % simple URL typesetting
\usepackage{booktabs}       % professional-quality tables
\usepackage{amsfonts}       % blackboard math symbols
\usepackage{nicefrac}       % compact symbols for 1/2, etc.
\usepackage{microtype}      % microtypography
\usepackage{lipsum}
\usepackage{fancyhdr}       % header
\usepackage{graphicx}       % graphics
\usepackage{float}
\graphicspath{{media/}}     % organize your images and other figures under media/ folder
\usepackage{siunitx} %正确显示浓度值并包括正确的单位
\usepackage{wrapfig}

\usepackage{markdown}
\usepackage{tabularx}
\usepackage{xcolor}
\usepackage{hyperref}
\usepackage{multirow}
\usepackage{listings}
\usepackage{longtable}
\usepackage{enumitem}
\hypersetup{
    colorlinks=true, % 设置为true表示链接带颜色，设为false则表示链接以框的形式呈现
    linkcolor=blue, % 内部链接（例如目录到章节的链接）的颜色设置为蓝色
    filecolor=magenta, % 文件链接的颜色
    urlcolor=cyan, % 外部URL链接的颜色
    citecolor=green % 引用链接的颜色
}

\definecolor{lightgray}{gray}{0.9}
\title{\raisebox{-20pt}{\includegraphics[width=0.1\textwidth]{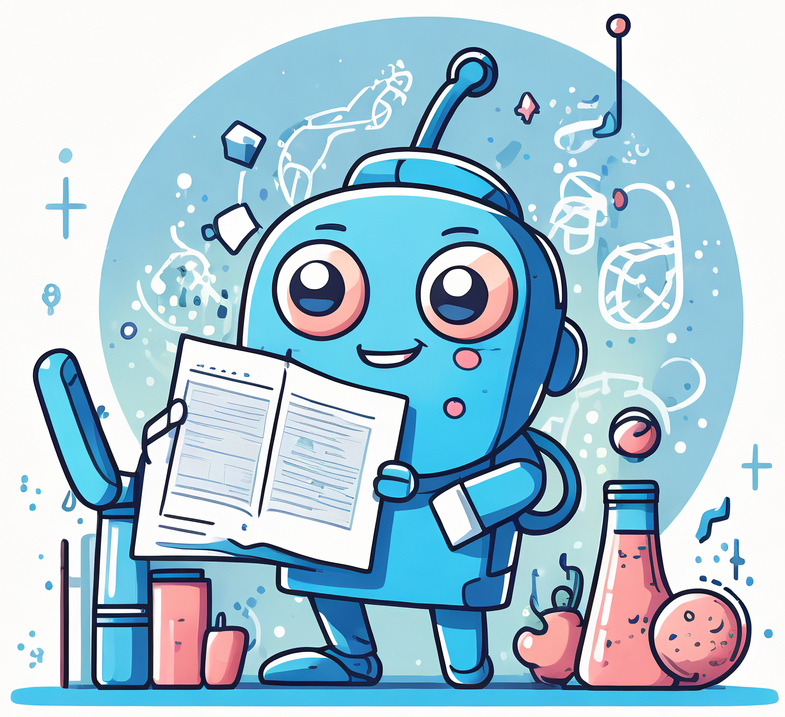}} Uni-SMART: Universal Science Multimodal Analysis and Research Transformer}

\author{%
\parbox{\linewidth}{\centering%
\textbf{Hengxing Cai$^{1}$}, \textbf{Xiaochen Cai$^{1}$}, \textbf{Shuwen Yang$^1$}, \textbf{Jiankun Wang$^1$}, \textbf{Lin Yao$^1$}, \textbf{Zhifeng Gao$^1$}, \\[0.5ex]
\textbf{Junhan Chang$^{1}$}, \textbf{Sihang Li$^1$}, \textbf{Mingjun Xu$^1$}, \textbf{Changxin Wang$^1$}, 
\textbf{Hongshuai Wang$^1$}, \\[0.5ex]
\textbf{Yongge Li$^1$}, \textbf{Mujie Lin$^1$}, \textbf{Yaqi Li$^1$}, \textbf{Yuqi Yin$^1$}, \textbf{Zheng Cheng$^2$}, \textbf{Zifeng Zhao$^2$}, \textbf{Linfeng Zhang$^{1,2}$} and \textbf{Guolin Ke$^{1}$} \\[1ex]
$^1$DP Technology \quad $^2$AI for Science Institute, Beijing\\[1ex]
\texttt{\{caihengxing, caixiaochen, yangsw, wangjiankun, yaol, gaozf, 
changjh, lisihang, xumj, wangchangxin, wanghongshuai, liyongge,
linmujie, liyq, yinyuqi, zhanglf, kegl\}@dp.tech}
}%\\
\\ \texttt{\{chengz, zhaozf\}@aisi.ac.cn}
}

\begin{document}

\maketitle

\vspace{-20pt}
\begin{center}
\href{http://uni-smart.dp.tech/}{http://uni-smart.dp.tech/}
\end{center}

\vspace{20pt}

\begin{abstract}

In scientific research and its application, scientific literature analysis is crucial as it allows researchers to build on the work of others. However, the fast growth of scientific knowledge has led to a massive increase in scholarly articles, making in-depth literature analysis increasingly challenging and time-consuming. 
The emergence of Large Language Models (LLMs) has offered a new way to address this challenge. Known for their strong abilities in summarizing texts, LLMs are seen as a potential tool to improve the analysis of scientific literature. However, existing LLMs have their own limits. Scientific literature often includes a wide range of multimodal elements, such as tables, charts, and molecule, which are hard for text-focused LLMs to understand and analyze. This issue points to the urgent need for new solutions that can fully understand and analyze multimodal content in scientific literature. To answer this demand, we present \textbf{Uni-SMART} (\underline{Uni}versal \underline{S}cience \underline{M}ultimodal \underline{A}nalysis and \underline{R}esearch \underline{T}ransformer), an innovative model designed for in-depth understanding of multimodal scientific literature. Through rigorous quantitative evaluation across several domains, Uni-SMART demonstrates superior performance over other text-focused LLMs. Furthermore, our exploration extends to practical applications, including patent infringement detection and nuanced analysis of charts. These applications not only highlight Uni-SMART's adaptability but also its potential to revolutionize how we interact with scientific literature. 

\end{abstract}

% By setting new benchmarks and revealing its practical benefits, Uni-SMART stands at the forefront of advancing multimodal understanding in scientific research, paving the way for more informed discoveries and innovations.

% keywords can be removed
% \keywords{First keyword \and Second keyword \and More}

\section{Introduction}
% 科研文献阅读是一个重要但耗时的工作
Scientific literature, encompassing patents and academic papers, constitutes a rich science data resource, including but not limited to drug properties and activities, reaction pathways, and manufacturing processes. 
However, extracting target information from this extensive corpus is a laborious and time-intensive task. 
It necessitates meticulous manual review, analysis, and extraction -- processes that are inherently slow and prone to human error \cite{hong2021challenges, nasar2018information}.
% 现有的非启发式的数据库仅提供了有限的检索能力，用户仍然要阅读分析检索到的文件才能得到答案
To enhance the efficiency of information retrieval, specialized databases like Sci-Finder \cite{scifinder} and Reaxys \cite{reaxys} have been developed.
However, their utility is constrained to document retrieval for molecule and reaction queries, lacking the capabilities of information extraction and knowledge comprehension to function as domain assistants.
Consequently, users must still engage in the tedious tasks of reading and analyzing the retrieved documents to extract definitive answers.
This limitation poses a significant bottleneck in the utilization of scientific data, hindering research progress and the rapid application of discoveries.
% 需要一个智能化的检索问答系统
Thus, researchers and practitioners require an intelligent navigator that can swiftly guide through the complexities of the latest scientific data, identify relevant information with precision, and present it in a digestible format.

\setlength{\intextsep}{20pt}
\setlength{\columnsep}{20pt}
\begin{wrapfigure}{r}{0.53\textwidth} 
  \label{fig:overview}
  \centering
  \includegraphics[width=0.5\textwidth]{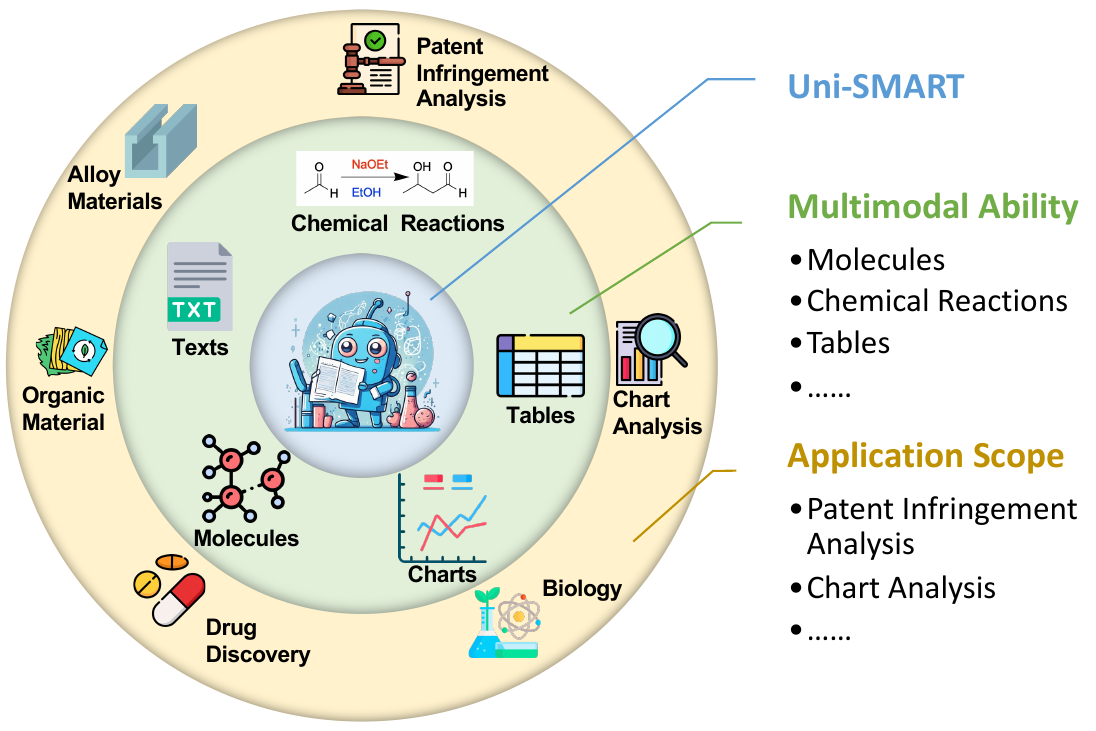} 
  \caption{Uni-SMART overview. It interprets multimodal data (\textit{e.g.,} texts, molecule structures, chemical reactions, charts, and tables), thereby facilitating a broad range of applications such as patent infringement analysis, chart analysis, and more.}
  \vspace{-12pt}
\end{wrapfigure}

% llm的出现在纯文本领域加速了检索问答
The emergence of Large Language Models (LLMs), represented by LLaMA \cite{llama}, Gemini \cite{gemini}, and GPT \cite{gpt3, rlhf, gpt4}, has marked a significant milestone in the evolution of natural language processing. 
These models have revolutionized the extraction of textual information from documents, enabling direct responses to queries using the extracted content. 
% 科学文献中有大量的多模态数据非常重要，但现有llm无法准确识别理解
Despite their proficiency, current LLMs are primarily designed for text extraction and often struggle with the multimodal aspects inherent in scientific literature, which include a large number of tables, charts, and reactions.
The extraction and interpretation of such multimodal data require an understanding that beyond texts and delves into visual and structural contents. 

% 引出uni-smart，简述其多模态能力
To address these challenges, we developed \textbf{Uni-SMART} (\underline{Uni}versal \underline{S}cience \underline{M}ultimodal \underline{A}nalysis and \underline{R}esearch \underline{T}ransformer), which extends the capabilities of LLMs beyond text, allowing for the interpretation of the multimodal content that is crucial in scientific literature. 
As illustrated in Figure \ref{fig:overview}, it is designed to recognize and analyze multimodal data, such as molecule structures, chemical reactions, charts, and tables, alongside textual content, facilitating a comprehensive understanding of scientific literature.
Such ability not only augments automated and precise information extraction but also enriches the interaction between researchers and the vast expanse of scientific knowledge. 

% 简述section3的evaluation
To rigorously assess the multimodal capabilities of Uni-SMART, a comparative analysis was conducted against several LLMs, such as GPT-4o, Gemini, and Claude3. 
Our assessment targets several data types critical to the comprehension of scientific documents: tables, charts, molecular structures, and chemical reactions.
The results demonstrate Uni-SMART's superior performance in all tested areas, especially in understanding and analyzing complex multimodal contents, thus highlighting its potential as a helpful assistant for scientific literature analysis.

% 简介报告outlines
In the following sections, Section \ref{sec:method} details Uni-SMART's data sources and iterative training approach.
Subsequently, Section \ref{sec:eval_new} -- Evaluation, presents detailed comparisons of Uni-SMART with several LLMs across a variety of modalities, showcasing its advanced capabilities in multimodal data interpretation. 
Then, Section \ref{sec:application} -- Application showcases some specific applications of Uni-SMART in the scientific domain.
Finally, Section \ref{sec:conclusion} discusses Uni-SMART's limitations and future research avenues, along with its potential impact on scientific research and technological advancement.
%Advanced techniques specifically designed for the integration of multimodal data are employed for fine-tuning, significantly enhancing its ability to interpret complex scientific information.

\section{Method}
\label{sec:method}
% This report primarily focuses on the capability assessment of Uni-SMART, a multimodal Transformer-style architecture developed by DP Technology for the analysis of scientific literature. 

% Subsequently, these data are transformed into multimodal formats suitable for Uni-SMART processing, including molecular structures, chemical equations, plain texts, mathematical expressions, charts, and tables, thereby facilitating nuanced understanding. 
% It processes, encodes, and fuses the input data, to achieve an in-depth comprehension of multimodal inputs. 

% Uni-SMART benefits from a hybrid training scheme that integrates conventional model training with human-in-the-loop alignment techniques.

% Uni-SMART leverages a broad spectrum of scientific data for training, including texts, tables, charts, molecular structures, and chemical reactions sourced from both public and proprietary scientific databases. 
% As depicted in Figure \ref{fig:smt_framework}, we source data from global patents, news articles, scientific publications, and market reports. 

\begin{figure}[t]
  \centering
  \includegraphics[width=\columnwidth]{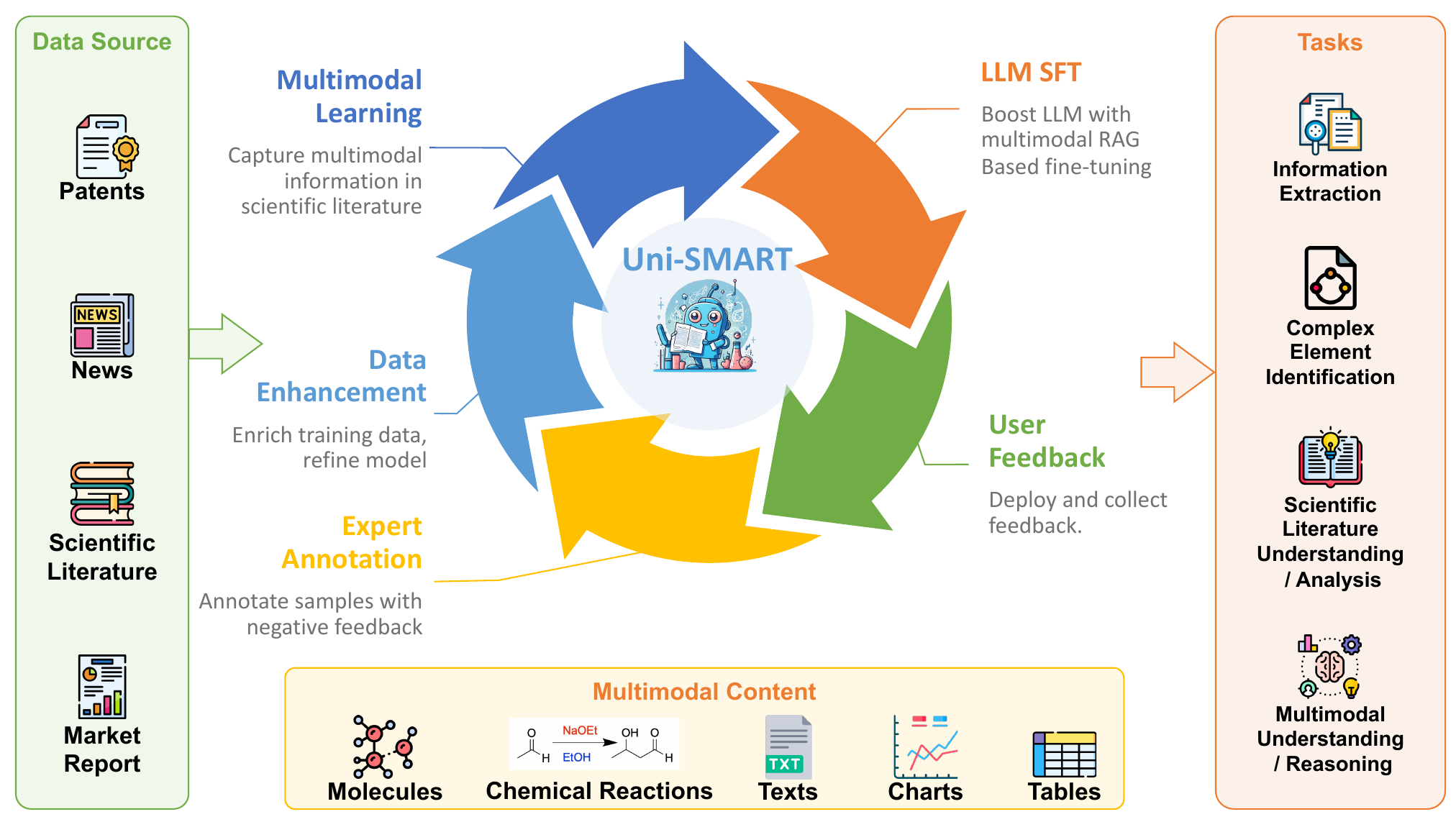}
  \caption{Uni-SMART model architecture.}
  \label{fig:smt_framework}
\end{figure}

As depicted in Figure \ref{fig:smt_framework}, Uni-SMART sources training data from a wide range of scientific literature from global patents, news articles, scientific publications, and market reports.
In particular, It adopts a cyclical, iterative approach to enhance its multimodal understanding capabilities, comprising the following key components:
\begin{itemize}[leftmargin=*]

    \item \textbf{Multimodal Learning:} During the initialization phase, the parsing model is trained with a limited set of multimodal data to recognize and extract diverse information elements from scientific literature. The output is formatted in a custom text format, similar to html format, to effectively represent multimodal elements.

    % \item \textbf{LLM Continual Pre-Training:} High-quality scientific data, such as textbooks and journals, are collected and parsed by the parsing model. The parsed data are then used for LLM  continual pre-training, enhancing its domain knowledge.
    
    \item \textbf{LLM SFT:} A series of valuable queries are constructed, particularly in multimodal scenarios. With Multimodal Retrieval-Augmented Generation (RAG), relevant content is recalled from the literature based on queries. Answers are built based on the queries and retrieved content. The query-answer data is then used to fine-tune the LLM. This process helps the LLM adapt to the custom input format and improves its ability to follow instructions.
    
    \item \textbf{User Feedback:} The parsing model and SFT-enhanced LLM are deployed in real applications, facilitating the collection of user feedback. 
    Samples receiving positive feedback are subsequently filtered and incorporated into the data enhancement, while those with negative feedback are subject to expert annotation before being integrated into the data enhancement process.

    \item \textbf{Expert Annotation:} Samples with negative feedback are carefully annotated by human experts. This step ensures that these models learns from their mistakes, with semi-automated tools assisting in this process to enhance efficiency. 
    Negative feedback cases typically fall into the following categories:
    \begin{enumerate}
        \item Multimodal element recognition errors;
        \item Recall content errors;
        % \item Domain knowledge errors
        \item Poor instruction-following.
    \end{enumerate}
    Detailed analysis of these error types facilitates targeted improvements.
    
    \item \textbf{Data Enhancement:} Finally, the annotated data, along with partial samples that received positive feedback, are added to the training dataset for data enhancement. The pipeline is optimized based on different types of negative feedback:
    \begin{enumerate}
        \item Multimodal element recognition errors: expand the training data for the parsing model;
        \item Recall content errors: optimize the Multimodal RAG scheme;
        % \item Domain knowledge errors: Strengthen the continued pre-training process.
        \item Poor instruction-following: enlarge and refine the dataset used for LLM SFT.
    \end{enumerate}
\end{itemize}

This iterative process is repeated to continually optimize Uni-SMART's overall performance. And significantly enhances Uni-SMART's performance in a variety of challenging tasks, such as information extraction, complex element identification, scientific literature understanding/analysis, and multimodal understanding/reasoning.

\section{Evaluation}
\label{sec:eval_new}

In this section, we perform a detailed quantitative evaluation on the capabilities of Uni-SMART and various available LLMs across modalities.
Table \ref{tab:statistical-overview-modal} presents the statistics of the benchmarks. 
For details of the specific evaluation tasks, please refer to SciAssess \cite{cai2024sciassess}. 

For our experiments, different methods are employed depending on the source of the models. Closed-source models are accessed via API calls, while open-source models are obtained from Hugging Face, deployed, and tested. Evaluation tasks that require an article as context involve converting the PDF content to text for input into the LLMs. If an LLM offers a built-in PDF parsing interface, we utilize this feature; otherwise, PyPDF2 is used for PDF parsing. Notably, Uni-SMART can directly read PDF files, allowing us to upload the original documents and pose questions directly to the model.
Due to input length limitations of the LLMs, tasks that require the full text of an article as context are executed in a zero-shot manner, whereas tasks that do not require such extensive context are tested using few-shot approaches.

The final results of these experiments are systematically presented in Table \ref{tab:all_result}, illustrating the performance of each model across the evaluated tasks.

\begin{table}[t]
\centering
\caption{Statistics of evaluation datasets}
\tiny
\setlength{\tabcolsep}{3pt}
\begin{tabular}{>{\centering\arraybackslash}m{1.5cm} >{\centering\arraybackslash}m{2.24cm} >{\centering\arraybackslash}m{3cm} >{\centering\arraybackslash}m{2cm} >{\centering\arraybackslash}m{2cm} >{\centering\arraybackslash}m{1cm}}
\toprule
Modality & Domain & Task & Question Type & Metric & \# Question \\
\midrule
\multirow{8}{1.5cm}{\centering Table} & Alloy Materials & Composition Extraction & Table Extraction & Table Recall & 244 \\
\noalign{\vskip 1pt}
\cline{2-6}
\noalign{\vskip 1pt}
& Drug Discovery & Affinity Extraction & Table Extraction & Table Recall & 40 \\
\noalign{\vskip 1pt}
\cline{2-6}
\noalign{\vskip 1pt}
& \multirow{4}{2cm}{\centering Organic Materials} & Electrolyte Table QA & Multi Choices & Accuracy & 100 \\
% \cmidrule{3-6}
& & OLED Property Extraction & Table Extraction & Table Recall & 13 \\
% \cmidrule{3-6}
& & Polymer Property Extraction & Table Extraction & Table Recall & 109 \\
% \cmidrule{3-6}
& & Solubility Extraction & Table Extraction & Table Recall & 100 \\
\midrule
\multirow{5}{1.5cm}{\centering Chart} & Alloy Materials & Alloy Chart QA & Multi Choices & Accuracy & 15 \\
\noalign{\vskip 1pt}
\cline{2-6}
\noalign{\vskip 1pt}
& Biology & Biology Chart QA & Multi Choices & Accuracy & 99 \\
\noalign{\vskip 1pt}
\cline{2-6}
\noalign{\vskip 1pt}
& Drug Discovery & Drug Chart QA & Multi Choices & Accuracy & 15 \\
\noalign{\vskip 1pt}
\cline{2-6}
\noalign{\vskip 1pt}
& Organic Materials & Polymer Chart QA & Multi Choices & Accuracy & 15 \\
\midrule
\multirow{6}{1.5cm}{\centering Molecule} & \multirow{4}{2cm}{\centering Drug Discovery} & Affinity Data Extraction & Table Extraction & Table Recall & 40 \\
% \cmidrule{3-6}
& & Tag to Molecule & Molecule Generation & Molecule Similarity & 50 \\
% \cmidrule{3-6}
& & Markush to Molecule & Molecule Generation & Molecule Similarity & 37 \\
% \cmidrule{3-6}
& & Molecule in Document & YES/NO & Accuracy & 50 \\
\noalign{\vskip 1pt}
\cline{2-6}
\noalign{\vskip 1pt}
& Organic Materials & OLED Property Extraction & Table Extraction & Table Recall & 13 \\
\midrule
\multirow{2}{1.5cm}{\centering Reaction} & Drug Discovery & Reaction QA & Multi Choices & Accuracy & 95 \\
% \cmidrule{3-6}
% & & ORD-standard Reaction Extraction & \\
\noalign{\vskip 1pt}
\cline{2-6}
\noalign{\vskip 1pt}
& Organic Materials & Reaction Mechanism QA & Multi Choices & Accuracy & 22 \\

\bottomrule
\end{tabular}
\label{tab:statistical-overview-modal}
\end{table}

\begin{table}[t]
\centering
\caption{Performance Comparison of LLMs on Diverse Tasks.}
\label{tab:all_result}
\tiny
\setlength{\tabcolsep}{2pt}
\begin{tabular}{>{
\centering\arraybackslash}m{1.2cm} | >{
\centering\arraybackslash}p{2.75cm} | >
{\centering\arraybackslash}p{1.2cm} >{
\centering\arraybackslash}p{0.75cm} >{
\centering\arraybackslash}p{0.75cm} >{
\centering\arraybackslash}p{0.75cm} >{
\centering\arraybackslash}p{0.75cm} >{
\centering\arraybackslash}p{0.75cm} >{
\centering\arraybackslash}p{0.75cm} >{
\centering\arraybackslash}p{0.75cm} >{
\centering\arraybackslash}p{0.75cm} >{
\centering\arraybackslash}p{0.75cm} >{
\centering\arraybackslash}p{0.75cm} >{
\centering\arraybackslash}p{1.25cm}
}
\toprule
Modality & Task & Uni-SMART & GPT-4o & GPT-4 & GPT-3.5 & Moonshot & Claude3 & Doubao & Gemini & Llama3 & DeepSeek & Qwen2 & Command $\text{R}^{\tiny{\text{+}}}$ \\
\midrule
\multirow{6}{1.2cm}{\centering Table}
& Composition Extraction & \textbf{0.511}  & 0.484 & 0.458 & 0.112 & 0.127 & 0.495 & 0.304 & 0.239 & 0.212 & 0.389 & 0.423 & 0.128 \\
& Affinity Extraction & \textbf{0.200}  & 0.072 & 0.042 & 0.025 & 0.040 & 0.097 & 0.050 & 0.040 & 0.064 & 0.017 & 0.075 & 0.043 \\
& Electrolyte Table QA & 0.850  & \textbf{0.890} & 0.790 & 0.370 & 0.670 & 0.870 & 0.710 & 0.880 & 0.460 & 0.720 & 0.620 & 0.450 \\
& OLED Property Extraction & \textbf{0.490}  & 0.336 & 0.406 & 0.201 & 0.037 & 0.477 & 0.259 & 0.093 & 0.263 & 0.292 & 0.392 & 0.234 \\
& Polymer Property Extraction & \textbf{0.705}  & 0.692 & 0.681 & 0.329 & \textbf{0.705} & 0.629 & 0.514 & 0.606 & 0.536 & 0.652 & 0.636 & 0.171 \\
& Solubility Extraction & \textbf{0.468}  & 0.435 & 0.440 & 0.410 & 0.363 & 0.426 & 0.371 & 0.397 & 0.399 & 0.432 & 0.400 & 0.351 \\
\midrule
\multirow{4}{1.2cm}{\centering Chart}
& Alloy Chart QA & \textbf{0.933}   & 0.533 & 0.600 & 0.333 & 0.333 & 0.400 & 0.467 & 0.667 & 0.467 & 0.333 & 0.400 & 0.200 \\
& Biology Chart QA & \textbf{0.616}  & 0.580 & 0.480 & 0.390 & 0.545 & 0.505 & 0.480 & \textbf{0.616} & 0.520 & 0.545 & 0.515 & 0.535 \\
& Drug Chart QA & \textbf{0.600}  & 0.333 & 0.400 & 0.067 & 0.400 & 0.200 & 0.533 & 0.533 & 0.400 & 0.400 & 0.400 & 0.533 \\
& Polymer Chart QA & \textbf{0.933} & 0.800 & 0.667 & 0.400 & 0.800 & 0.467 & 0.867 & 0.800 & 0.867 & 0.733 & \textbf{0.933} & 0.800 \\
\midrule
\multirow{5}{1.2cm}{\centering Molecule}
& Affinity Extraction & \textbf{0.200}  & 0.072 & 0.042 & 0.025 & 0.040 & 0.097 & 0.050 & 0.040 & 0.064 & 0.017 & 0.075 & 0.043 \\
& Tag to Molecule & \textbf{0.188}  & 0.040 & 0.022 & 0.000 & 0.016 & 0.035 & 0.094 & 0.169 & 0.034 & 0.014 & 0.000 & 0.031 \\
& Markush to Molecule & \textbf{0.686}  & 0.642 & 0.654 & 0.431 & 0.504 & 0.675 & 0.239 & 0.526 & 0.491 & 0.470 & 0.379 & 0.376 \\
& Molecule in Document & \textbf{0.720}  & 0.580 & 0.700 & 0.500 & 0.460 & 0.480 & 0.560 & 0.640 & 0.680 & 0.460 & 0.460 & 0.460 \\
& OLED Property Extraction & \textbf{0.490}  & 0.336 & 0.406 & 0.201 & 0.037 & 0.477 & 0.259 & 0.093 & 0.263 & 0.292 & 0.392 & 0.234 \\
\midrule
\multirow{2}{1.2cm}{\centering Reaction}
& Reaction QA & \textbf{0.768}   & 0.705 & 0.674 & 0.442 & 0.253 & 0.663 & 0.442 & 0.305 & 0.611 & 0.368 & 0.442 & 0.316 \\
& Reaction Mechanism QA & 0.682  & 0.545 & 0.636 & 0.455 & 0.545 & 0.455 & 0.636 & \textbf{0.727} & 0.500 & 0.545 & 0.591 & 0.591 \\
\bottomrule
\end{tabular}
\end{table}

\subsection{Table}
Tables play a pivotal role in scientific literature, presenting complex data and findings in a highly structured manner and thereby contributing significantly to scientific discovery and the dissemination of knowledge \cite{desai2021tablex}. They facilitate the intuitive display of experimental data and enable the efficient summarization and comparison of research outcomes, becoming an indispensable component of scientific investigation. Consequently, enhancing the capability to understand tables is crucial for the automated processing and analysis of scientific documents. The utilization of table data spans a wide array of applications, such as trend analysis, which can reveal developmental trajectories in research fields, and comparative studies, which can elucidate differences in experimental outcomes under varying research methodologies or conditions \cite{milosevic2019framework}.

To assess the table understanding capabilities of our model, Uni-SMART, compared to other
LLMs, we designed a diverse set of tasks across different domains. These tasks were specifically tailored to evaluate how well each model could interpret and extract information from tables.

In our thorough assessment, the Uni-SMART showcased its exceptional proficiency in understanding and extracting table data from the scientific literature.
Among six assessed tasks, it surpassed other models in five and delivered a competitive performance in the remainder.
Notably, in the "Composition Extraction" and "Solubility Extraction" tasks, Uni-SMART achieved "Table Recall" of 0.511 and 0.468, respectively, significantly outperforming its counterparts.
The objectives of these two tasks are to extract necessary information from tables within articles and organize it into a specified format. These results indicate that Uni-SMART excels in handling and understanding tabular data, particularly in information extraction and formatting tasks.

However, in the Electrolyte Table QA task, Uni-SMART's performance was slightly below the state-of-the-art model (GPT-4o). In this task, LLMs are required to answer questions about details in the table with the question type of multiple-choice, indicating that Uni-SMART still has room for improvement in understanding the details within tables.

Moreover, all models under-performed in the "Affinity Extraction" task, indicating the high complexity of these tasks.
This task requires LLMs to organize molecules and corresponding experimental result from articles. It necessitates the model's ability to integrate molecules with table information and demands strong long-context understanding capabilities. To better address these issues in the future, a possible direction is to enhance LLMs' abilities in long context comprehension and information matching.

\subsection{Chart}
Charts are essential tools in scientific literature, offering a visual representation of data that can significantly enhance the comprehension and communication of complex information. By succinctly illustrating trends, comparisons, and patterns, charts enable researchers to convey their findings more effectively and intuitively \cite{kallimani2013extraction}. Therefore, the ability to accurately interpret and analyze charts is vital for the automated processing and understanding of scientific documents. 

To assess the chart interpretation capabilities of Uni-SMART, we conducted a series of ChartQA tasks spanning various scientific domains. These tasks rigorously evaluated models' ability to analyze and clarify data presented in charts, with a particular focus on identifying trends and extracting meaningful insights. 

\begin{figure}[t]
  \centering
  \includegraphics[width=\columnwidth]{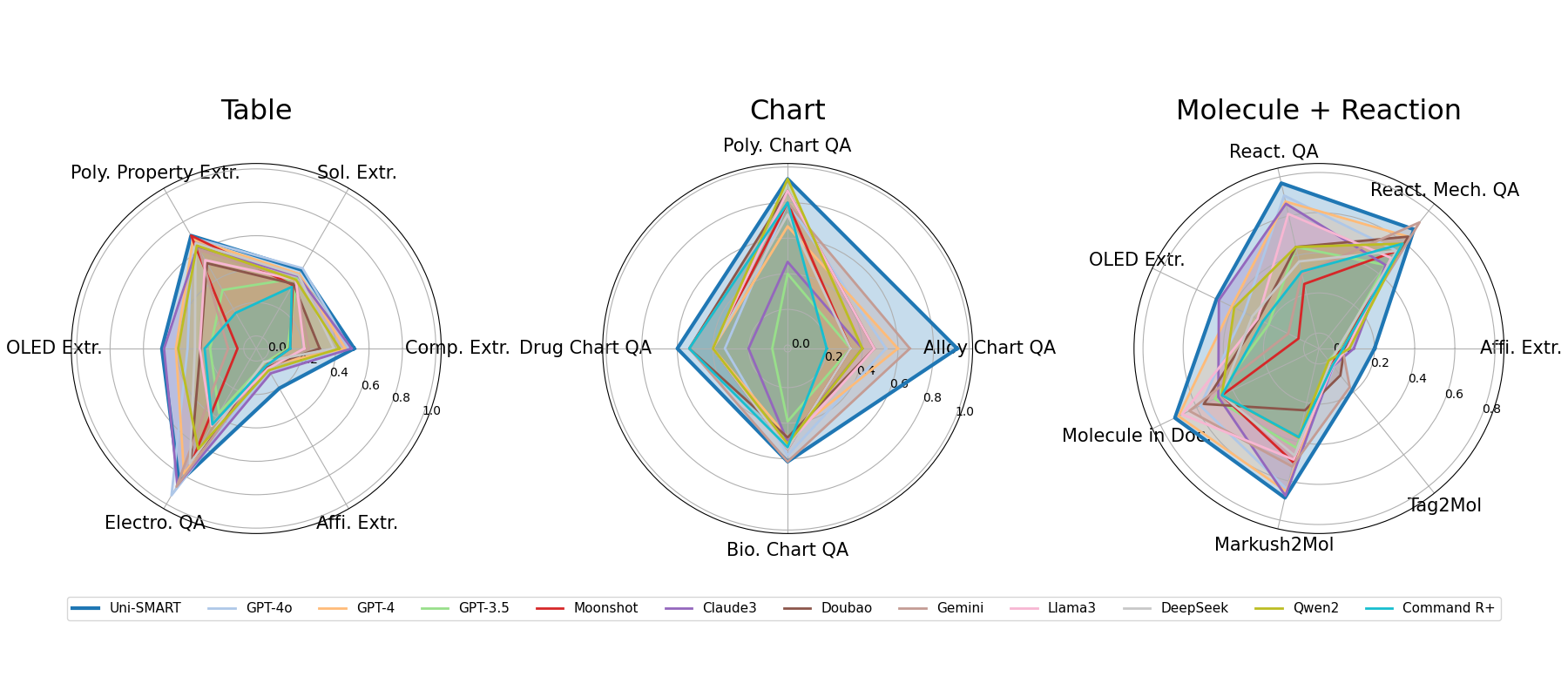}
  \caption{Comparative Performance of LLMs Across Different Modalities.}
  \label{fig:radar_result}
\end{figure}

Analysis of the results presented in Figure \ref{fig:radar_result} indicates that Uni-SMART significantly outperforms existing models in Chart QA tasks across diverse scientific domains, particularly excelling in the tasks of Alloy Chart QA and Drug Chart QA, with leading scores of 0.933 and 0.600, respectively.
However, both in the Biology Chart QA, where the Gemini model scored 0.616, and in the Polymer Chart QA task, where Qwen2 also achieved a top score of 0.933—matching Uni-SMART's performance in both cases—it is evident that these models have strong capabilities in analyzing scientific charts.
These results underscore the intense competition among different models in the context of chart understanding. The future direction for development likely focuses on enhancing accuracy and applicability across various scientific domains.

% However, despite Uni-SMART's leading position, its advantage narrows in the task of Biology Chart QA, where Gemini also achieves equally high scores. 
% This could be because the task is relatively simpler, and LLMs might be able to achieve the correct answers based solely on text rather than a deep understanding of the chart's content.

% 分子结构
\subsection{Molecule}
In scientific literature, molecule or molecular structures hold fundamental importance, especially within fields such as chemistry, pharmacology, and biology. 
Comprehending molecular structures is crucial for analyzing research outcomes, predicting chemical behaviors, and innovating novel compounds \cite{blaschke2002information, swain2016chemdataextractor}. 
Deciphering the molecular structure of a newly discovered drug compound, for instance, can provide essential insights into its therapeutic potential and biological interactions, which is crucial for enhancing our understanding of related scientific literature.

To evaluate our model's ability to understand molecular structures, we carried out a series of tasks involving molecules, polymers, and Markush structures, which are common in chemistry and medicine studies.
These tasks were designed to assess the model's capability in deciphering information from representations of molecular structures.

The Uni-SMART model showed outstanding performance in molecular structure understanding tasks, particularly in processing complex molecular structures within the scientific literature.
For example, in the "Molecule in Document" task, Uni-SMART achieved an accuracy of 0.720, significantly outperforming other models. This task requires LLMs to determine whether a molecular structure has appeared in the literature, demonstrating Uni-SMART's strong capabilities in accurate molecule identification and information retrieval.

Such achievements highlight Uni-SMART's considerable advantage in parsing molecular structure, potentially attributed to its access to richer, more specialized training data sources, effective preprocessing of molecular structure, and architecture designed specifically for multimodal data. 
Despite these strengths, there remains room for improvement in Uni-SMART's comprehension of molecular structure. 
Future enhancements may include strengthening the model's understanding of molecule details and expanding the training dataset to encompass a broader range of molecule types. 
These efforts are expected to further augment Uni-SMART's applicability in reading and interpreting scientific literature, especially those involving molecular structure.

%化学反应
\subsection{Chemical Reactions}
Understanding chemical reactions is crucial in scientific literature, particularly within chemistry and its related disciplines \cite{guo2021automated}. 
The ability to accurately parse and analyze the details of chemical reactions enhances readers' grasp of complex material, enabling a deeper understanding of experimental results and theoretical discussions.

To assess Uni-SMART's capability to comprehend chemical reactions, we designed two tasks aimed at challenging the model's proficiency in interpreting chemical reactions. 
These tasks focus on understanding the reactants, products, and conditions of chemical reactions, as well as grasping the underlying mechanisms and clarifying the significance of these reactions within a broader scientific context.

Uni-SMART exhibited outstanding performance in "Reaction QA" task, achieving an accuracy of 0.768.
However, in the "Reaction Mechanism QA" task, Uni-SMART's performance was slightly below that of the Gemini. 
This task requires LLMs to accurately parse and interpret the step-by-step mechanisms of chemical reactions from the article, indicating that Uni-SMART still has room for improvement in understanding complex reactions.

\section{Application}
\label{sec:application}

In this section, we showcase how Uni-SMART significantly improves the efficiency and quality of scientific research. 
Through several selected case studies, we will present the specific applications of Uni-SMART in the scientific domain. 
These cases allow us to qualitatively assess the model's capabilities in addressing various research challenges. 
For more details and updates, please refer to the Uni-SMART website at \url{http://uni-smart.dp.tech/}.

% In this section, we present a selection of qualitative cases to demonstrate the Uni-SMART's proficiency in cross-modal understanding.
% In the realm of scientific literature analysis, the capability to synthesize information across multiple modalities is of paramount importance. Scientific documents often contain diverse modalities of information, such as text, tables, charts, and molecules, each contributing complementary insights that collectively elucidate comprehensive scientific concepts or research findings. Thus, an efficacious scientific literature understanding system necessitates not merely the processing of unimodal information but also the adept integration of multimodal data to facilitate a more profound and holistic comprehension. This chapter aims to qualitatively evaluate the proficiency of the Uni-SMART in the integrated understanding of multimodal information within scientific texts, by examining a series of illustrative cross-modal cases.

\subsection{Patent infringement analysis}
\begin{figure}[t]
  \centering
\includegraphics[width=0.9\columnwidth]{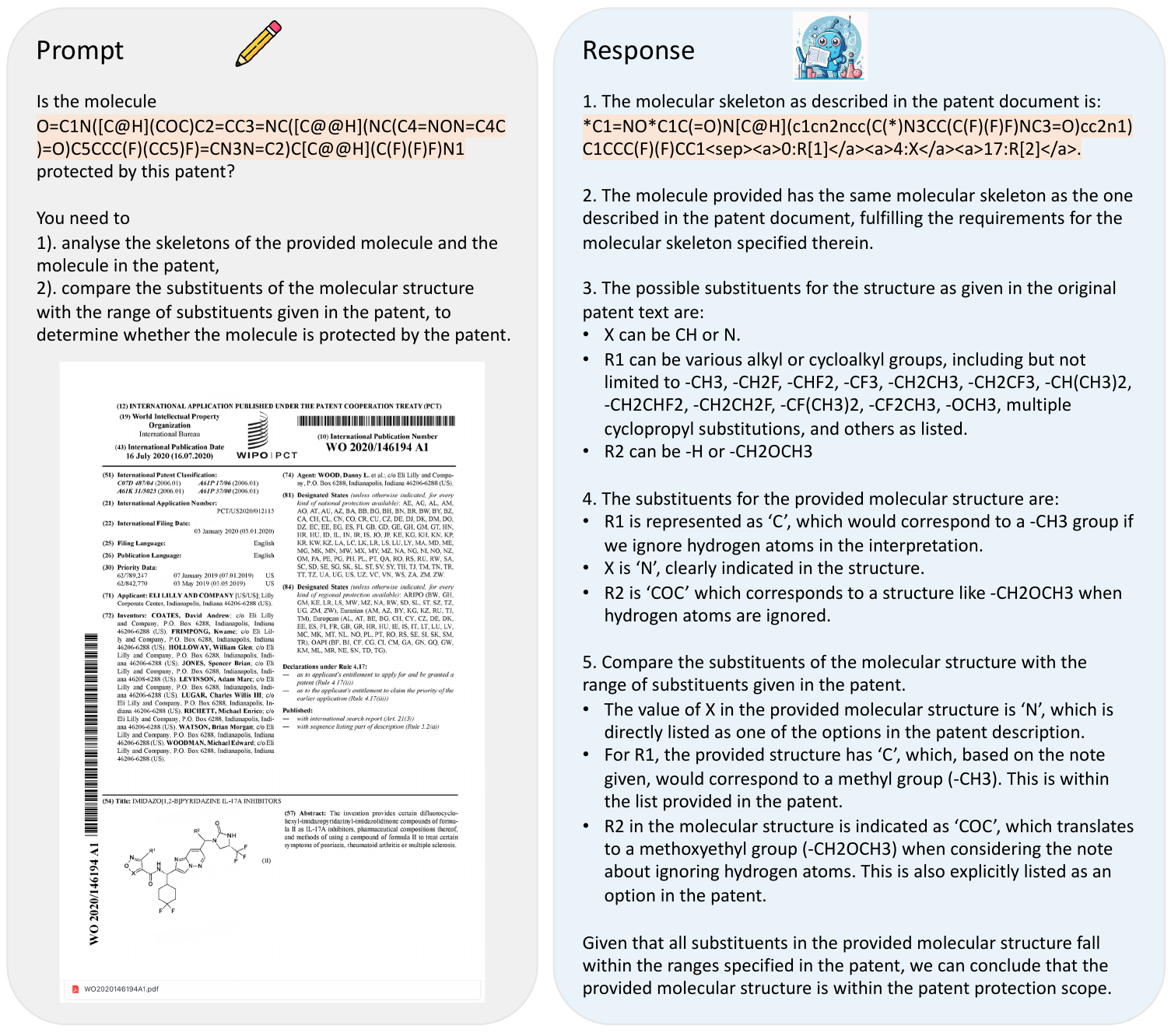}
  % \vspace{-5pt}
  \caption{Patent infringement analysis.}
  % \vspace{-20pt}
  \label{fig:patent-infringement}
\end{figure}

In today's research and industrial domains, the correct understanding and application of patent information have become increasingly crucial, especially in the fields of chemistry and drug development \cite{melethil2005patent}. Accurately determining whether a compound is protected by existing patents is essential for avoiding potential patent infringements. However, this task presents numerous technical challenges, particularly the need for precise analysis of complex chemical structures and their comparison with descriptions in patent documents. This requires not only a deep understanding of chemical structures but also the ability to process and analyze information from different modalities, such as chemical structure and patent text descriptions.

In this case study, we utilize Uni-SMART to analyze whether a specific chemical structure is protected by a particular patent. The process involves receiving the patent document and compound uploaded by the user, analyzing the molecular skeleton and substituents of the compound, and then comparing this information with the range given in the patent document.

\begin{figure}[t]
\centering
% \vspace{-10pt}
\includegraphics[width=0.95\columnwidth]{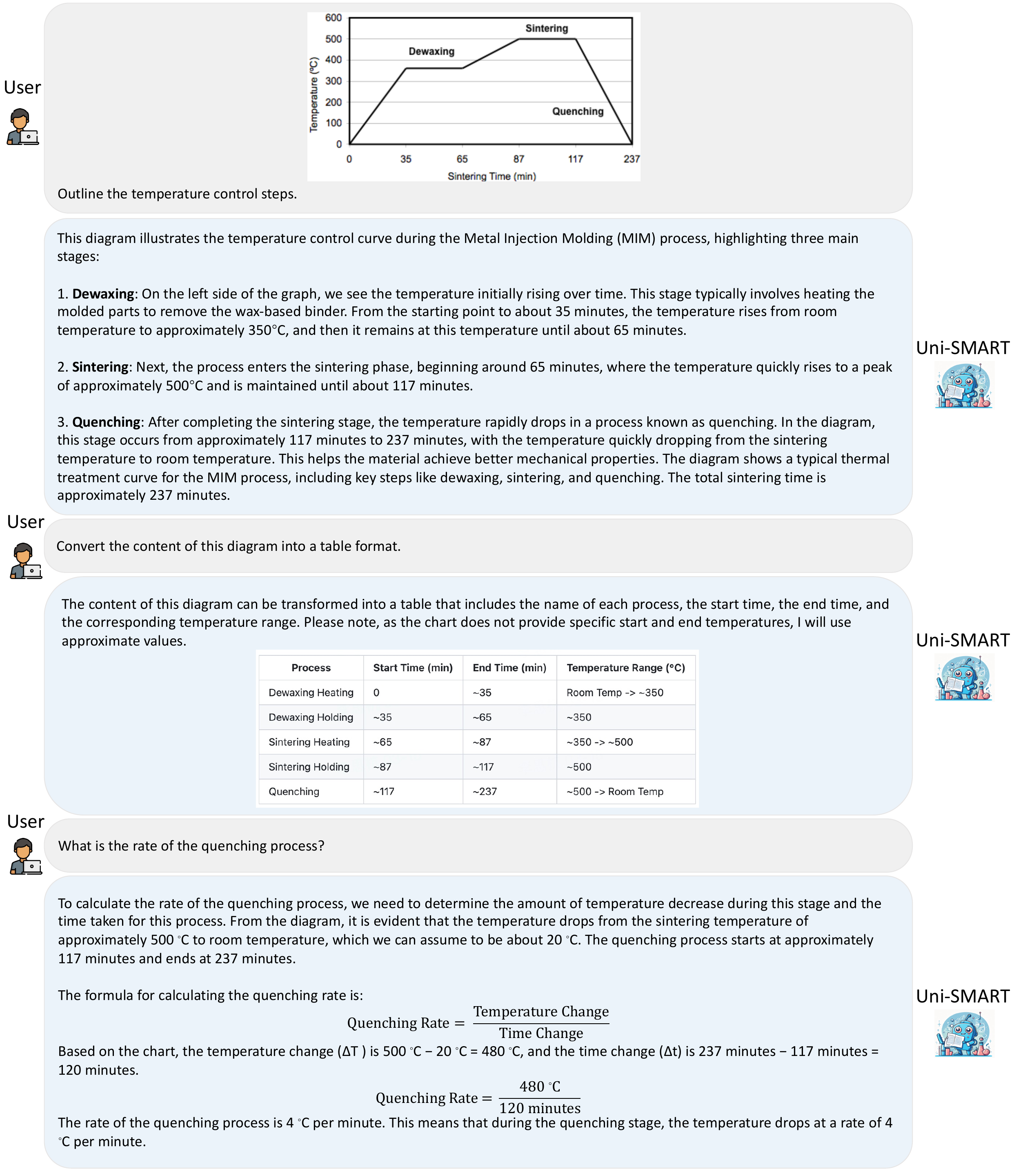}
  % \vspace{-6pt}
  \caption{Understanding the Metal Injection Molding (MIM) Process.}
  \label{fig:MIM}
\end{figure}

As shown in Figure \ref{fig:patent-infringement}, Uni-SMART analyzes the provided compound structure and accurately determines whether the compound falls within the patent protection scope by considering multiple factors, such as the molecular skeleton and substituents. This process not only demonstrates Uni-SMART's efficient handling of cross-modal information between chemical structure and patent text but also highlights its strong capability in understanding and applying patent information. Applying Uni-SMART in practical scenarios like patent infringement protection can help users avoid potential patent risks, which is of significant importance for advancing scientific research and industrial development.

\subsection{Chart analysis}

In scientific literature, charts serve as a crucial tool for conveying complex data and experimental results. They play an essential role in the deep understanding of technical details by presenting data trends in an intuitive form. However, the effective understanding and analysis of charts in scientific literature encounter significant challenges. These include but are not limited to, the simultaneous parsing of visual elements and textual descriptions within charts, the accurate identification of data points, understanding the scientific processes behind the data, and performing subsequent complex calculations and logical reasoning.

In this case study, Uni-SMART was utilized to analyze a chart showing the temperature control curve during the Metal Injection Molding (MIM) process.
As illustrated in Figure \ref{fig:MIM}, Uni-SMART accurately described the temperature control curve of the MIM process, precisely identifying the key data points in the chart. Following user instructions, it successfully converted this information into a tabular format. Furthermore, the model demonstrated its powerful capability in performing mathematical calculations and logical reasoning, especially in calculating the quenching rate.

\section{Discussion and Conclusion}
\label{sec:conclusion}

% 总结、重申uni-smart在科研文献场景下，多模态理解方面的优势。我们进行了定量的评估，结果如何。背后归功于数据/模型
In this report, we introduce Uni-SMART for in-depth understanding of multimodal information within the scientific literature. Through rigorous quantitative evaluation, Uni-SMART demonstrates significant performance gain in interpreting and analyzing multimodal contents in scientific documents, such as tables, charts, molecular structures, and chemical reactions, compared with other competitors. 
The success of Uni-SMART lies in its innovative cyclic iterative process that continuously refines its multimodal understanding capabilities, leveraging a robust dataset and a combination of multimodal learning, supervised fine-tuning, user feedback, expert annotation, and data enhancement to achieve superior performance in scientific literature analysis.
% The effectiveness of Uni-SMART is attributed to the innovative multimodal Transformer architecture, which underwent extensive pre-training on scientific literature data rich in multimodal content and carefully designed fine-tuning. 

% 除了定量的评估，令我们更兴奋的是一系列使用场景。
Beyond quantitative assessment, we are particularly excited about Uni-SMART's potential to address scientific challenges through practical applications.
From patent infringement analysis to complex material science chart interpretation, Uni-SMART's cross-modal understanding capabilities offer new perspectives and tools for research and technological development, showcasing its potential to facilitate research processes and accelerate discovery phases.

% 尽管...，目前还是有一些局限性
Despite Uni-SMART's strong ability in multimodal scientific literature understanding, we acknowledge that there is still room for improvement. 
This includes enhancing the model's understanding of highly complex and specialized content, as well as reducing hallucinations.
We believe that through continuous research and development, these limitations will be addressed, making Uni-SMART an even more powerful and flexible tool for scientific research assistance.

% 强调这个研究对于未来科研工作、技术发展的影响，以及潜在商业应用。
In summary, the research and development of Uni-SMART mark a significant advancement in the field of multimodal scientific literature understanding. 
By providing scientists and researchers with an efficient tool for deep understanding and analysis of scientific documents, Uni-SMART not only facilitates the accumulation and innovation of scientific knowledge but also paves the way for future scientific work, technological development, and potential commercial applications. 
As we continue to improve and expand Uni-SMART, we look forward to its greater role in promoting scientific discovery and technological innovation.

\newpage

%Bibliography
\bibliographystyle{unsrt}  
\bibliography{references}  

\appendix
% \newpage
% \section{Appendix}

% \newpage
% \appendix
% \section*{\centering \huge Appendix}
% \section{Experiment Setup}
% \label{app:exp_setup}

% We use API calls for closed-source models, while open-source models are obtained from HuggingFace, deployed, and tested.
% For tasks requiring an article as context, the PDF content is converted to text and input into the LLM.
% If the LLM offers built-in PDF parsing interface, we use it; otherwise, we use PyPDF2 for PDF parsing.
% Uni-SMART, capable of directly reading PDF files, allows us to upload the original PDF files and pose questions directly to the model.
% Due to input length limitations of the LLMs, tasks that require the full text of an article as context are executed in a zero-shot manner, whereas tasks that do not require such extensive context are tested using few-shot approaches.

\end{document}